\documentclass{article}
\usepackage{spconf,amsmath,graphicx}
\usepackage{bm}
\usepackage{times}
\usepackage{latexsym}
\usepackage{tablefootnote}
\usepackage{graphicx}
\usepackage{multirow}
\usepackage{multicol}
\usepackage{booktabs}
\usepackage{url}
\usepackage{footmisc} 
\usepackage{hyperref}
\usepackage{footnote}
\usepackage{amsfonts,amssymb}

\hypersetup{hidelinks}

\title{Type-aware Decoding via Explicitly Aggregating Event Information for Document-level Event Extraction}
\name{Gang Zhao, Yidong Shi, Shudong Lu, Xinjie Yang, Guanting Dong, Jian Xu, Xiaocheng Gong, Si Li\sthanks{Corresponding author.}}
\address{School of Artificial Intelligence, Beijing University of Posts and Telecommunications, China}
\begin{document}
\ninept
\maketitle
\begin{abstract}
Document-level event extraction~(DEE) faces two main challenges: arguments-scattering and multi-event. 
Although previous methods attempt to address these challenges, they overlook the interference of event-unrelated sentences during event detection and neglect the mutual interference of different event roles during argument extraction. 
Therefore, this paper proposes a novel Schema-based Explicitly Aggregating~(SEA) model to address these limitations. 
SEA aggregates event information into event type and role representations, enabling the decoding of event records based on specific type-aware representations. 
By detecting each event based on its event type representation, SEA mitigates the interference caused by event-unrelated information.
Furthermore, SEA extracts arguments for each role based on its role-aware representations, reducing mutual interference between different roles. 
Experimental results on the ChFinAnn and DuEE-fin datasets show that SEA outperforms the SOTA methods.
\end{abstract}
\section{Introduction}
\label{sec:intro}

Document-level Event Extraction~(DEE) aims to detect events and extract event arguments of pre-defined types from documents, which plays important roles in various fields, such as question answering~\cite{han-etal-2021-ester}, financial analysis, speech understanding, etc.
In contrast to sentence-level event extraction~\cite{lin2020joint,TANL,lu2021text2event,hsu2022degree,lu-etal-2022-unified} which extracts events within a sentence, DEE faces two specific challenges: arguments-scattering and multi-event.
As shown in Figure \hyperlink{1}{1}, arguments-scattering refers to the situation where arguments of an event record are dispersed across multiple sentences that are far apart from each other.
On the other hand, multi-event signifies that a document can contain several correlated records that share common arguments.

Despite the recent advancements~\cite{zheng2019doc2edag,xu-etal-2021-git,yang-etal-2021-document,liang2022raat} in tackling aforementioned challenges, the existing approaches tend to adhere to a common paradigm. These methods typically involve encoding a document into sentence and entity representations, and subsequently decoding event records solely using these representations.
However, such a DEE paradigm exhibits several limitations.
Firstly, they detect each type of event using all sentence representations, thereby overlooking the interference caused by event-unrelated sentences or sentences related to other event types.
% Secondly, they extract arguments of different event roles based on the same entity representations, rendering them susceptible to mutual interference, particularly when an entity plays multiple event roles.
Secondly, the argument extraction for different event roles relies on the same set of entity representations, rendering them susceptible to mutual interference, particularly when an entity plays multiple roles.

\begin{figure}[t]
\hypertarget{1}
\centering
\includegraphics[scale=0.935]{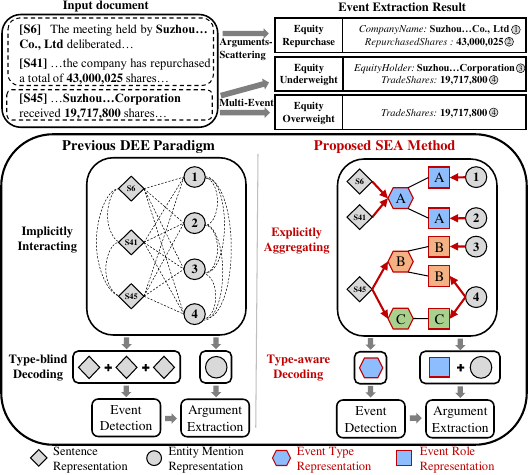}
\vspace{-0.4cm}
\caption{
A simplified example from the ChFinAnn dataset, showcasing a comparison between the previous methods and our approach.
}
\vspace{-0.4cm}
\label{fig:my_label}
\end{figure}

In this paper, to address the limitations of the previous paradigm, we propose a Schema-based Explicitly Aggregating model~(SEA) for DEE.
As depicted in Figure \hyperlink{1}{1}, the key idea of SEA is to aggregate event information into corresponding event type and role representations, enabling the decoding of event records based on specific type-aware event representations.
By detecting each type of event based on its event type representation, SEA effectively mitigates interference from other events and event-unrelated information.
Furthermore, SEA extracts arguments for each role based on its role-aware entity representations, thus alleviating the mutual interference among different roles.
In detail, SEA begins by creating event queries for various event types and roles in the event schema.
Then, SEA utilizes the proposed Event Representation Extractor to obtain event representations that are aware of the global document context.
Subsequently, the event type information is aggregated into event type nodes, while the argument information is aggregated into role nodes within the proposed Explicitly Aggregating Graph Network.
Finally, SEA detects events based on the event type representations and extracts arguments using role-aware entity representations, which are enhanced by incorporating role representations.

Our contributions are summarized as follows:

1)~To alleviate the interference suffered by the previous DEE paradigm, we propose the SEA model which explicitly aggregates event information into event representations and decodes event records using specific type-aware representations.

2)~Experimental results on widely used datasets ChFinAnn and DuEE-fin demonstrate that SEA outperforms the state-of-the-art methods, effectively tackles DEE challenges, and alleviates the limitations of the previous DEE paradigm.

\section{Related Works}
Most previous works of Event Extraction~\cite{liu-etal-2018-jointly, yan2019event, du-cardie-2020-event,lu2021text2event,hsu2022degree} concentrate on Sentence-level Event Extraction~(SEE) based on the ACE 2005~\cite{walker2006ace} dataset. 
SEE methods typically begin by detecting event trigger words and then extracting arguments within a sentence.
However, SEE methods face challenges in extracting event records that span multiple sentences, a scenario frequently encountered in real-world situations.
Therefore, there has been increasing interest in Document-level Event Extraction~(DEE) recently.

Recent studies on DEE~\cite{yang-etal-2018-dcfee, zheng2019doc2edag, yang-etal-2021-document, xu-etal-2021-git,liang2022raat} are mainly based on the financial document-level datasets, ChFinAnn and DuEE-fin, and focus on the two critical challenges: arguments-scattering and multi-event.
\cite{yang-etal-2018-dcfee} extracts events at sentence level first, then identifies the key events and pads the missing arguments from the neighboring sentences.
\cite{zheng2019doc2edag} obtains document-aware sentence and entity representations via multi-level Transformers, and treats DEE as a directed acyclic graph generation.
% \cite{zheng2019doc2edag} acquires document-aware sentence and entity representations using multi-level Transformers and formulates DEE as a directed acyclic graph generation problem.
\cite{yang-etal-2021-document} proposes a multi-granularity decoder to decode event records in a  parallel manner based on sentence and entity representations.
\cite{xu-etal-2021-git} utilizes a heterogeneous graph to enhance sentence and entity representations and tracks multiple records with a global memory.
% \cite{liang2022raat} proposes a tailored transformer structure to incorporate relation information between arguments and enhance the extraction.
\cite{liang2022raat} proposes a tailored transformer structure to incorporate relation information between arguments.

In general, recent works on DEE mainly focus on obtaining better sentence and entity representations, or improving the decoding mechanisms based on them.
However, their type-blind decoding paradigm overlooks the interference of event-unrelated sentences during event detection, and neglects the mutual interference of different roles during argument extraction.
In contrast to these works, we propose to model event information explicitly via event representations and decode event records based on specific type-aware representations to alleviate the interference.
% To the best of our knowledge, we are the first work to perform type-aware decoding via explicitly aggregating event information, which enhances the performance and tackles the critical challenges of DEE effectively.

\section{Methodology}
\label{sec:method}
Figure \hyperlink{2}{2} shows the architecture of SEA, which comprises the following key components: Document Encoder, Event Representation Extractor~(ERE), Explicitly Aggregating Graph Network~(EAGN), and Type-aware Event Record Decoder.
For a clearer exposition, we first clarify certain terminologies:
1)~\emph{Entity Mention} refers to a text span within a document that pertains to an entity object.
2)~\emph{Event Argument} represents an entity fulfilling a specific role in an event record. Event roles are predefined for each event type.
3)~\emph{Event Record} denotes an instance of a particular event type that includes arguments for various roles within the event.
4)~\emph{Event Schema} defines all the event types and event roles that appear in the dataset.

\subsection{Document Encoder}
Given a document $\mathcal{D}={\{s_i\}}^{|\mathcal{D}|}_{i=1}$, where $s_i$ is the $i^{th}$ sentence containing $|s_i|$ tokens, we utilize a Transformer~\cite{vaswani2017attention}  to encode $s_i$ following~\cite{zheng2019doc2edag,xu-etal-2021-git,yang-etal-2021-document}:
\setlength{\abovedisplayskip}{0.1cm}
\setlength{\belowdisplayskip}{0.1cm}
\begin{equation}
\widetilde{s}_i=Transformer\mbox{-}1{(s_i)}\in\mathbb{R}^{|s_i|\times d}
\end{equation}
Next, we obtain the sentence representation $\mathcal{S}_i \in \mathbb{R}^d$ by max-pooling $\widetilde{s}_i$ and adding the sentence position embedding: $\mathcal{S}_i=Max(\widetilde{s}_i)+PosEmb(s_i)$.
After that, we employ a CRF~\cite{lafferty2001conditional} layer to recognize entities as candidate event arguments and get the entity recognition loss $\mathcal{L}_{er}$:
\setlength{\abovedisplayskip}{0.1cm}
\setlength{\belowdisplayskip}{0.1cm}
\begin{equation}
\mathcal{L}_{er}=-\sum\limits_{i=1}^{|\mathcal{D}|}logP(\hat{y}_{s_i}|\widetilde{s_i})
\end{equation}
where $\hat{y}_{s_i}$ is the golden label sequence of $s_i$.
Thus, we can derive the entity mention $M_i=\{m_j\}_{j=1}^{|M_i|}$, where $m_j$ is the constituent token representation of $M_i$. Finally, we apply max-pooling on $M_i$ to get the entity mention representation $\mathcal{M}_i\in\mathbb{R}^d$.

\subsection{Event Representation Extractor}
We propose the ERE to create event type and role representations which are utilized to aggregate event information in EAGN.
The event schema defines all the event types and event roles that appear in a DEE dataset.
We begin by creating learnable embeddings, referred to as event queries, for each event type and event role: $\textbf{T}=\{T_m\}_{m=1}^{N_T}$, $\textbf{R}_m=\{R_{m_n}\}_{n=1}^{N_{R_m}}$, where $N_T$ is the number of event types, and $R_{m_n}$ denotes the role query for the $m^{th}$ event type.
Then, we concatenate $\widetilde{s}_i$ with event queries and employ a Transformer encoder to obtain sentence-aware event representations:
\setlength{\abovedisplayskip}{0.1cm}
\setlength{\belowdisplayskip}{0.1cm}
\begin{equation}
[\overline{s}_i,\overline{T}_m^i,\overline{\textbf{R}}_m^i]=Transformer\mbox{-}2([\widetilde{s}_i;T_m;\textbf{R}_m])
\end{equation}
Eventually, we derive document-aware event representations by max-pooling the sentence-aware event representations:
$\mathcal{T}_m=Max(\{\overline{T}_m^i\}_{i=1}^{|D|})\in\mathbb{R}^d$, $\mathcal{R}_{m_n}=Max(\{\overline{R}_{m_n}^i\}_{i=1}^{|D|})\in\mathbb{R}^d$, where $\mathcal{T}_m$ and $\mathcal{R}_{m_n}$ are event type and role representations.

\begin{figure}[t]
\hypertarget{2}
\centering
\includegraphics[scale=0.29]{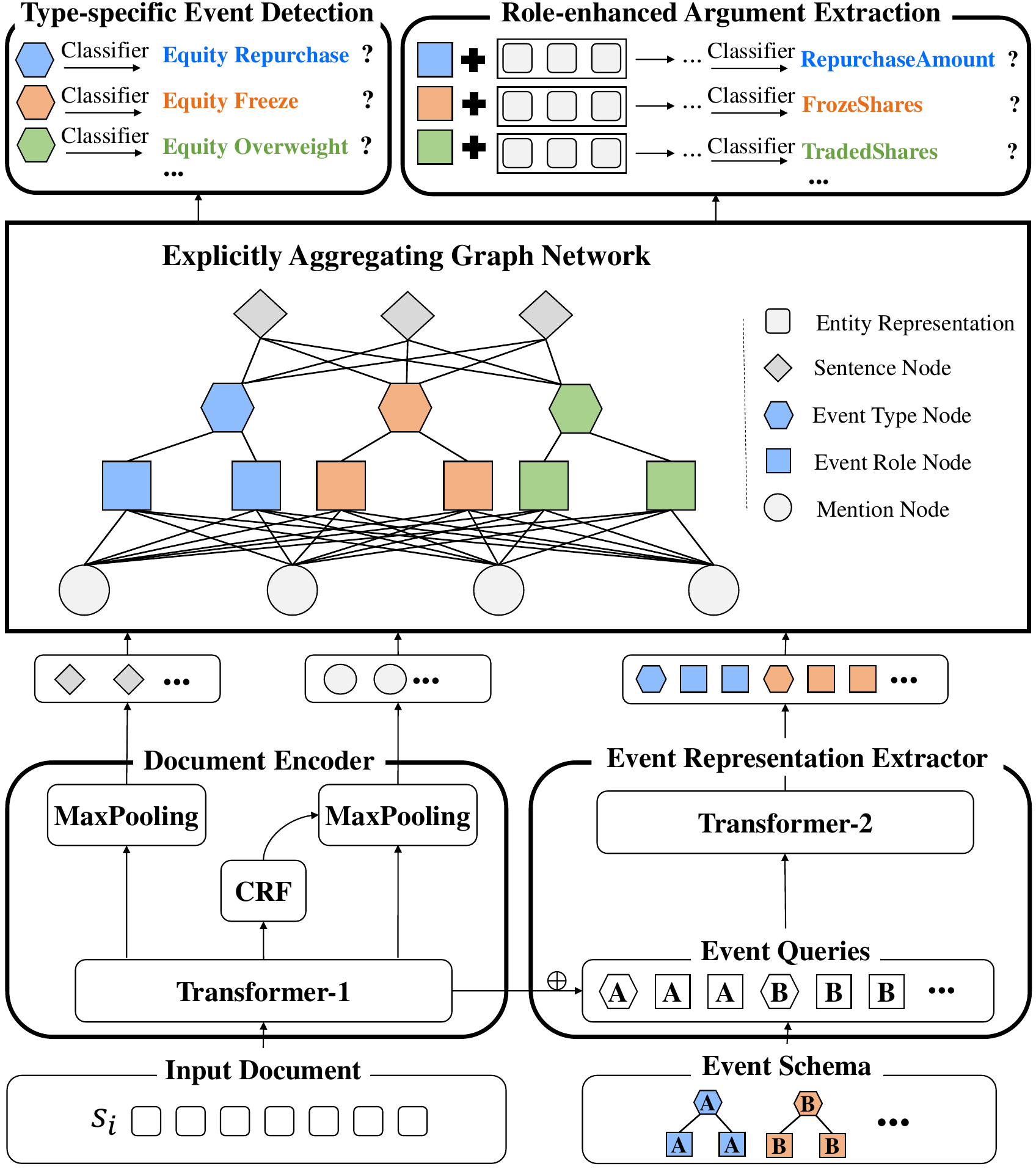}
\vspace{-0.4cm}
\caption{
The overall architecture of SEA. }
\vspace{-0.4cm}
\label{fig:my_label}
\end{figure}

\subsection{Explicitly Aggregating Graph Network}
The EAGN is proposed to aggregate event information into event representations.
As Figure \hyperlink{2}{2} shows, EAGN consists of four types of nodes: event type node $\mathcal{T}_m$, event role node $\mathcal{R}_{m_n}$, sentence node $\mathcal{S}_i$, and entity mention node $\mathcal{M}_i$.
We design the following types of edges to aggregate event information:

% 1)~We connect $\mathcal{S}_i$ with $\mathcal{T}_m$ to aggregate event trigger information into event type nodes.
% It allows us to detect each kind of event on respective $\mathcal{T}_m$.
1)~$\mathcal{S}_i$ is connected to $\mathcal{T}_m$ to aggregate event trigger information into event type nodes, enabling the detection of various events on respective $\mathcal{T}_m$ nodes.

2)~$\mathcal{M}_i$ is connected to $\mathcal{R}_{m_n}$ to aggregate argument information into role nodes, aiding in the extraction of corresponding arguments.

3)~Considering the inherent relation between an event type and its roles, $\mathcal{T}_m$ is connected to $\mathcal{R}_{m_n}$ to facilitate the information interaction among them.

% Following~\cite{xu-etal-2021-git}, we also connect different $\mathcal{S}_i$, $\mathcal{M}_i$ of the same entity, and $\mathcal{M}_i$ appearing in the same sentence to capture document-level contexts.
Following~\cite{xu-etal-2021-git}, connections are also established between different $\mathcal{S}_i$, $\mathcal{M}_i$ associated with the same entity, and $\mathcal{M}_i$ appearing in the same sentence to capture document-level contexts.
Finally, a Graph Convolutional Network~\cite{kipf2016semi} is employed to derive information-aggregated event representations ($\widetilde{\mathcal{T}}_m$,  $\widetilde{\mathcal{R}}_{m_n}$), as well as document-aware representations ($\widetilde{\mathcal{S}}_i$, $\widetilde{\mathcal{M}}_i$).
The aggregation is learned under the supervision of type-aware decoding.

\hypertarget{20}{\subsection{Type-aware Event Record Decoder}}
To mitigate the interference from different event types and roles, we perform type-aware decoding based on the event representations.

\textbf{Type-specific Event Detection.}
To alleviate the interference of sentences that are event-unrelated or related to different types of events, we treat event detection as a binary classification on each event type $\widetilde{\mathcal{T}}_m$ instead of $\{\widetilde{\mathcal{S}}_i\}_{i=1}^{|D|}$ in previous works:
\setlength{\abovedisplayskip}{0.1cm}
\setlength{\belowdisplayskip}{0.1cm}
\begin{equation}
C_m=Sigmoid(FFN(\widetilde{\mathcal{T}}_m))
\end{equation}
We optimize the classification loss $L_{ed}$ when training:
\setlength{\abovedisplayskip}{0.1cm}
\setlength{\belowdisplayskip}{0.1cm}
\begin{equation}
\mathcal{L}_{ed}=-\sum\limits_{m=1}^{|\bm{T}|}(f(\widehat{C}_m)logC_m +g(\widehat{C}_m)log(1-C_m))
\end{equation}
where $\widehat{C}_m$ is golden classification label.

\textbf{Role-enhanced Argument Extraction.}
Following~\cite{zheng2019doc2edag}, we extract arguments as a tree path expanding subtask.
However, to alleviate the interference among different roles, we enhance the entity representations by adding role representations and extract arguments for each role based on its specific role-aware entity representations:
\setlength{\abovedisplayskip}{0.1cm}
\setlength{\belowdisplayskip}{0.1cm}
\begin{equation}
A_{mn}=Sigmoid(FFN(\widetilde{\bm{E}}+\widetilde{\mathcal{R}}_{mn}))
\end{equation}
where $\widetilde{\bm{E}}=\{\widetilde{E}_i\}_{i=1}^{|\widetilde{\bm{E}}|}$ and $\widetilde{E}_i$ is the entity representation obtained by max-pooling the corresponding mention representations $\{\widetilde{\mathcal{M}}_j\}_{j=1}^{|\widetilde{E}_i|}$.
The argument extraction loss $\mathcal{L}_{ae}$ is as follows:
$$\mathcal{L}_{ae}=-\sum\limits_{m=1}^{|\bm{T}|}\sum\limits_{n=1}^{|\bm{R}_m|}\sum\limits_{i=1}^{|\widetilde{\bm{E}}|}(f(\widehat{A}_{mni})logA_{mni}$$
\setlength{\abovedisplayskip}{0.1cm}
\setlength{\belowdisplayskip}{0.1cm}
\begin{equation}
+g(\widehat{A}_{mni})log(1-A_{mni}))
\end{equation}
where $\widehat{A}_{mni}$ is golden label, $|\bm{T}|$ is the event type number, and $|\bm{R}_m|$ is the role number of the $m^{th}$ event type.
Finally, we sum the losses mentioned above as the final loss,  $\mathcal{L}=\lambda_1\mathcal{L}_{er}+\lambda_2\mathcal{L}_{ed}+\lambda_3\mathcal{L}_{ae}$.
% Details of $\mathcal{L}_{er}$, $\mathcal{L}_{ed}$ and $\mathcal{L}_{ae}$ are shown in Appendix \hyperlink{321}{B}.

\section{Experiments}
\subsection{Experimental Setup}
\hspace{1em}\textbf{Datasets.}
Following~\cite{liang2022raat}, we evaluate our method on two widely-used DEE datasets: ChFinAnn~\cite{zheng2019doc2edag} and DuEE-fin~\cite{DuEE-fin-2021}.
% ChFinAnn contains 32,040 documents and is associated with 5 event types and 35 role types.
% Recently proposed DuEE-fin encompasses 13 event types and 92 role types, posing a greater challenge as it involves more types of events and lacks auxiliary entity annotations.
ChFinAnn is an influential public financial DEE dataset proposed by~\cite{zheng2019doc2edag}, which consists of 32,040 documents focusing on 5 event types and 35 event roles.
We adopt the standard split for ChFinAnn as~\cite{zheng2019doc2edag}.
DuEE-fin~\cite{DuEE-fin-2021} is a recently released public financial DEE dataset comprising 13 event types and 92 event roles extracted from common financial events.
DuEE-fin is more challenging since it involves more types of events and lacks auxiliary entity annotations.
% We treat the argument annotation as the entity annotation for DuEE-fin in the experiments.
We divide DuEE-fin into train, development, and test sets as 5,258, 892, and 1,023 documents, respectively.
Data statistics are shown in Table \hyperlink{1000}{1}.
% Statistics of both datasets are shown in Table \hyperlink{1000}{1}.

\begin{table}[t]
\hypertarget{1000}
\centering
\footnotesize
\renewcommand\arraystretch{1.3}
\setlength\tabcolsep{2pt}%调列距

\begin{tabular}{c|ccc|c}
\hline
\multirow{2}{*}{\textbf{Subset}} & \multicolumn{3}{c|}{\bf{Example Number}}                          & \multirow{2}{*}{\begin{tabular}[c]{@{}c@{}} \bf{Associated Sentences} \\ \bf{(Average)} \end{tabular}} \\ \cline{2-4}
                                 & \multicolumn{1}{c|}{Full} & \multicolumn{1}{c|}{S.}   & M.   &                                                                    \\ \hline
Train                            & \multicolumn{1}{c|}{25632 / 5258} & \multicolumn{1}{c|}{18114 / 3511} & 7518 / 1747 & -                                             \\ \hline
Dev                              & \multicolumn{1}{c|}{3204 / 892}  & \multicolumn{1}{c|}{2207 / 599}  & 997 / 293  & -                                             \\ \hline
Test                             & \multicolumn{1}{c|}{3204 / 1023}  & \multicolumn{1}{c|}{2413 / 696}  & 791 / 327  & -                                             \\ \hline

I                                & \multicolumn{1}{c|}{1267 / 466}  & \multicolumn{1}{c|}{-}     & -    & 6.43 / 1.69                                          \\ \hline
II                               & \multicolumn{1}{c|}{1306 / 497}  & \multicolumn{1}{c|}{-}     & -    & 8.81 / 3.75                                          \\ \hline
III                              & \multicolumn{1}{c|}{631 / 60}   & \multicolumn{1}{c|}{-}     & -    & 14.29 / 8.66                                         \\ \hline
\end{tabular}
\vspace{-0.15cm}
\caption{
Data statistics of ChFinAnn/DuEE-fin.
% \textbf{Overall} shows Micro F1 scores~(\%) on the complete test set.
\textbf{S.} and \textbf{M.} indicate single-record and multi-record.
\textbf{I}, \textbf{II}, and \textbf{III} are subsets of test documents with increasing arguments-scattering sentences.
}
\vspace{-0.4cm}
\end{table}

\textbf{Baselines and Metrics.}
% We compare SEA with the following SOTA methods:
% DCFEE \cite{yang-etal-2018-dcfee}, Doc2EDAG \cite{zheng2019doc2edag}, DEPPN \cite{yang-etal-2021-document}, GIT \cite{xu-etal-2021-git}, and ReDEE \cite{liang2022raat}.
% DCFEE has two versions, DCFEE-O and DCFEE-M, while Doc2EDAG has a variant GreedyDec.
We compare our SEA with the following baseline methods:
1)~\textbf{DCFEE}~\cite{yang-etal-2018-dcfee}, which extracts events at the sentence level and pads the missing arguments from the neighboring sentences. 
DCFEE has two implementation versions, \textbf{DCFEE-O} and \textbf{DCFEE-M}.
2)~\textbf{Doc2EDAG}~\cite{zheng2019doc2edag}, which acquires document-aware representations via multi-level Transformers and decodes events as a directed acyclic graph construction.
Doc2EDAG has a variant \textbf{GreedyDec}, which decodes only one event record greedily.
3)~\textbf{DEPPN}~\cite{yang-etal-2021-document}, which employs a multi-granularity decoder to decode event records in a parallel manner.
4)~\textbf{GIT}~\cite{xu-etal-2021-git}, which uses a heterogeneous graph to interact sentence and entity representations, and tracks multiple records with a global memory.
5)~\textbf{ReDEE}~\cite{liang2022raat}, which employs an argument relation extraction subtask to incorporate relation information and enhance the extraction.
We evaluate all models with the role-level Micro F1 score.

\textbf{Implementation Details.}
In the implementation of SEA, we follow most of the hyperparameter settings in previous works and make little effort to select the best hyperparameters.
% Specifically, we set $\lambda_1$= 0.05,  $\lambda_2$=$\lambda_3$=0.95 for the overall loss function.
Following~\cite{zheng2019doc2edag}, we set the values of $\lambda_1$= 0.05, $\lambda_2$=$\lambda_3$=0.95, and use the schedule sampling strategy~\cite{NIPS2015_e995f98d} to alleviate the exposure bias in the entity recognition.
We utilize 2 layers Transformer encoder in the ERE module, and employ BERT pre-trained by~\cite{cui2020revisiting} as our document encoder following~\cite{liang2022raat}.
We use three layers GCN to aggregate event information and set the dropout rate to 0.1.
We set the learning rate to 5e-5, the batch size to 64, and choose Adam~\cite{kingma2014adam} as our optimizer.
We conduct experiments in this paper on Tesla-V100 GPUs, and report the average results of three runs.
All experiments of baseline models are conducted based on the released code in their original papers~\cite{zheng2019doc2edag,xu-etal-2021-git,yang-etal-2021-document,liang2022raat}.
% We report the average result of 3 runs for each experiment.
Since ReDEE~\cite{liang2022raat} has not released the detailed relation designs for DuEE-fin, we leave the comparison for future studies.
The results of DEPPN on the ChFinAnn dataset are evaluated using their released model checkpoint~\cite{yang-etal-2021-document}.
The code of SEA will be released soon for future studies.

% More details about the implementation, datasets, and baselines are shown in Appendix \hyperlink{322}{B}, \hyperlink{33}{C}, and \hyperlink{34}{D}.

% \textbf{Implementation Details.}
% We use 2 layers Transformer in ERE, and use 8 layers Transformer in Document Encoder for experiments on ChFinAnn.
% % We use 8 and 2 layers Transformer in the Document and Event Ontology Encoder of SEA for experiments on ChFinAnn.
% Considering DuEE-fin lacks entity annotations, we use BERT pre-trained by \cite{devlin2018bert} in Document Encoders of SEA and all baseline models for experiments on DuEE-fin.
% The complete experimental setting is shown in Appendix \hyperlink{6}{A}.

% Main hyper-parameters used in the experiments are shown below.
% We use 8 and 4 layers Transformer in encoding and decoding module respectively.
% The transformer’s hidden layers and feed-forward layers have the embedding size of 768 and 1024.
% During training, we set $\lambda_1$= 0.05, $\lambda_2$=$\lambda_3$=0.95 and $\gamma$= 3.
% We also use L= 3 layers of GCN, and set dropout rate to 0.1, batch size to 64.
% SEA chooses the Adam \cite{kingma2014adam} optimizer and set the learning rate as 1$\emph{e}^{-4}$, and the maximum epochs are set as 100.

\begin{savenotes}
\begin{table*}[t]
\hypertarget{11}
\centering
\small
\renewcommand\arraystretch{0.9}
\setlength\tabcolsep{8pt}%调列距
\label{tab:methodcompare}
\begin{tabular}{l|cc|ccc|c|cc|ccc|c}
\toprule
\multirow{2}{*}{\bf Model} & \multicolumn{6}{c}{\bf ChFinAnn}&\multicolumn{6}{|c}{\bf DuEE-fin}\\ 
%\cmidrule(lr){2-7} \cmidrule(lr){8-11} 
\cmidrule{2-13}
% \hline

{\bf } & {\bf S.} & {\bf M.} & {\bf I} & {\bf II} & {\bf III} & {\bf Overall}& {\bf S.} & {\bf M.} &{\bf I} & {\bf II} & {\bf III} & {\bf Overall} \\
\hline 
DCFEE-O & 66.2 & 51.0 & 68.5 & 57.1 & 51.0 & 59.1& 67.1 & 51.4 & 68.1&54.8&46.2 & 58.5\\
DCFEE-M & 60.6 & 49.3 & 62.5 & 55.0 & 46.9 & 55.2& 57.7 & 44.1 & 60.5&45.9&38.7 & 49.8\\
GreedyDec & 78.3 & 36.5 & 73.0 & 60.7 & 44.8 & 61.1& 73.0 & 48.1 & 68.9&56.5&46.5 & 59.8\\
Doc2EDAG & 84.0 & 67.4 & 83.2 & 75.7 & 69.1 & 76.3& 69.5 & 58.8 &70.6&62.3&49.4 & 63.5\\
GIT & 87.5 & 70.2 & 85.5 & 79.2 & 72.7 & 79.5& \underline{73.8} & \underline{64.6} & \underline{74.1} & \underline{67.4} & \underline{59.3} & \underline{68.7} \\
ReDEE & \underline{88.1} & \underline{73.4} & {\bf87.7} & \underline{81.0} & \underline{74.5} & \underline{81.5} & - & - & - & - & - & -\\

\hline 
{\bf SEA} &  {\bf 88.3} & {\bf 74.5} & \underline{ 87.3} & {\bf 81.8} & {\bf 75.6} & {\bf 82.2}& {\bf 76.9} & {\bf 68.7} & {\bf 76.9}& {\bf 71.7}& {\bf 61.6} & {\bf 72.4}\\
\hline 
w/o ERE & -0.4 & -1.0 & -0.6 &-1.0& -0.3& -0.8&-3.2&-1.3&-2.0&-0.8&-7.5&-2.1 \\
w/o EAGN & -1.4 & -1.5 & -0.1 &-2.3& -1.5& -1.5&-5.4&-5.1&-4.0&-4.9&-9.6&-5.1 \\
w/o EventType & -0.4 & -1.1 & -0.5 &-0.8& -1.0& -0.8&-2.2&-3.0&-6.5&-3.4&-4.6&-2.6  \\
w/o Role & -2.4& -0.7 & -1.2 &-1.9& -1.9& -1.8&-2.6&-2.3&-1.3&-1.6&-4.3&-1.7  \\

\bottomrule
\end{tabular}
\vspace{-0.2cm}
\caption{
Main experimental result on ChFinAnn and DuEE-fin datasets.
% \textbf{Overall} shows Micro F1 scores~(\%) on the complete test set.
\textbf{S.} and \textbf{M.} indicate Micro F1 scores~(\%) on single-record and multi-record test documents.
\textbf{I}, \textbf{II}, and \textbf{III} are F1 scores on test documents with increasing arguments-scattering severities.
% \emph{w/o} marks the variants of ablation study.
\textbf{Overall} indicates F1 scores on the full test set.
% Results marked by * are cited from \cite{xu-etal-2021-git}.
% Results below \textbf{SEA} are ablation studies
Models marked by \textbf{w/o} are variants of the ablation study. 
}
\vspace{-0.3cm}
\end{table*}
\end{savenotes}

\subsection{Results and Analysis}
As Table \hyperlink{11}{2} shows, SEA outperforms other baseline models on both ChFinAnn and DuEE-fin.
Specifically, SEA improves 0.7\% overall Micro F1 score on ChFinAnn and 3.7\%  F1 on DuEE-fin, compared with the SOTA methods, ReDEE and GIT.

\textbf{Multi-event Challenge.}
To study the influence of type-aware decoding and explicitly aggregating on multi-event extraction, we conduct experiments on test documents with single~(S.) and multiple~(M.) records separately.
Compared to SOTA, SEA surpasses 0.2/3.1\% Micro F1 on single-record and 1.1/4.1\% on multi-record test documents of ChFinAnn/DuEE-fin.
% The improvement brought by SEA is mainly in multi-record documents, suggesting that our method tackles the multi-event challenge effectively.
The primary improvement brought by SEA is observed on multi-record documents, indicating the effectiveness of our method in addressing the multi-event challenge.
% This is because SEA separates mixed event information into respective event representations and extracts each kind of event on its specific event representations.
It is achieved by separating mixed event information into respective event representations and decoding each type of events utilizing type-aware representations.

\begin{figure}[t]
\hypertarget{3}
\centering
\includegraphics[scale=0.43]{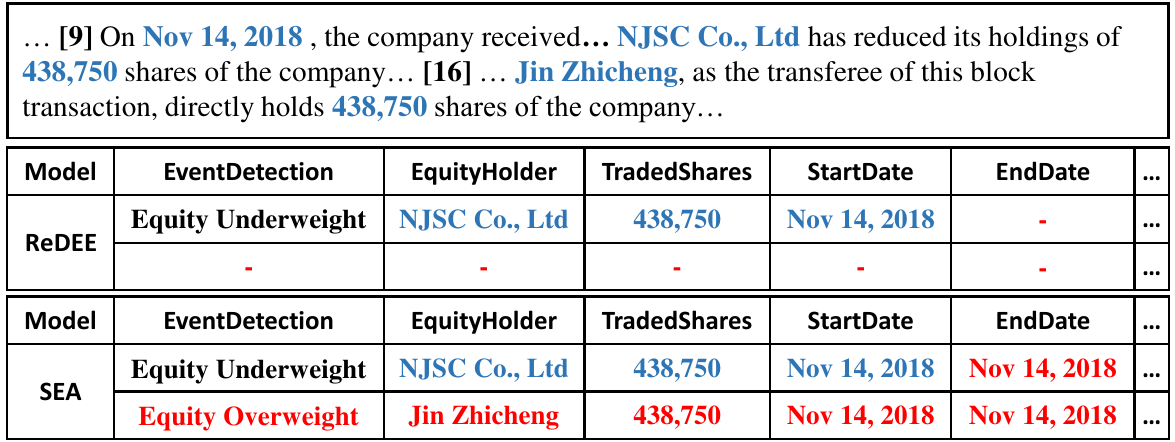}
\vspace{-0.6cm}
\caption{
Case study of the proposed SEA and ReDEE, with their key predicting differences colored in red.
% Related entities are colored in blue.
}
\vspace{-0.4cm}
\label{fig:my_label}
\end{figure}
\textbf{Arguments-scattering Challenge.}
To verify the effectiveness of our approach on arguments-scattering extraction, we divide the test set into three subsets I/II/III based on the increasing number of record-associated sentences.
The Records in set III pose the greatest challenge since their arguments scatter across most sentences.% Compared to SOTA, SEA improves -0.4/0.8/1.1\% on I/II/III of ChFinAnn and 2.8/4.3/2.3\% on DuEE-fin, showing that SEA addresses the arguments-scattering challenge effectively. %这里在I的指标是下降的，要不要改一下说法
% Compared to SOTA, SEA improves 0.8/1.1\% on II/III of ChFinAnn and 2.8/4.3/2.3\% on DuEE-fin, showing that SEA addresses the arguments-scattering challenge effectively.
Compared to SOTA, SEA improves 0.8/1.1\% F1 on II/III of ChFinAnn and 4.3/2.3\% F1 on DuEE-fin, showing that SEA addresses the arguments-scattering challenge effectively.
It is because SEA concentrates scattered event information into event representations by explicitly aggregating event information.

\textbf{Ablation Study.}
To verify the key components of SEA, we perform ablation tests on 4 variants:
% 1)~\emph{-ERE}, removing ERE and using event queries to aggregate event information.
% 2)~\emph{-EAGN}, removing EAGN and omitting event information aggregation.
% 3)~\emph{-EventType}, detecting event based on sentence representations as previous methods.
% 4)~\emph{-Role}, extracting argument without role-aware representations as previous methods.
1)~\emph{-ERE} removes the ERE and uses the event queries to aggregate event information.
2)~\emph{-EAGN} removes the EAGN and omits the event information aggregation.
3)~\emph{-EventType} detects event based on sentence representations as previous methods.
4)~\emph{-Role} extracts argument without role-aware representations as previous methods.
From Table \hyperlink{11}{2}, we can observe that:
1)~Obtaining document-aware event representations via ERE is important and contributes 0.8/2.1\% overall F1-score.
2)~Aggregating event information is essential and enhances 1.5/5.1\% overall F1-score.
3)~Detecting event based on specific event type representation improves 0.8/2.6\% overall F1, especially 1.1/3.0\% F1 on multi-event extraction.
4)~Extracting argument using role-aware representations is critical which improves 1.8/1.7\% overall F1, and 1.9/4.3\% F1 in the most challenging arguments-scattering situation.

\begin{figure}[t]
\hypertarget{4}
\centering
\includegraphics[scale=0.98]{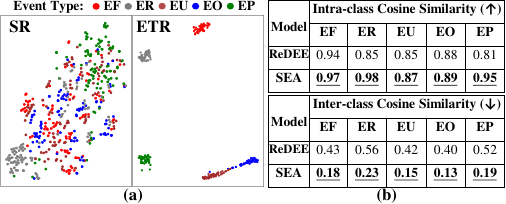}
\vspace{-0.4cm}
\caption{
% Distribution of ETR and record-related SR of \emph{Equity Freeze~(EF)} records.
% ETR: Event Type Representations, SR: Sentence Representations.
(a)~Visualization of SR and ETR.
(b)~Averaged intra/inter-class cosine similarities on each event type. 
}
\vspace{-0.4cm}
\label{fig:my_label}
\end{figure}

\textbf{Case Study.}  Figure \hyperlink{3}{3} shows a representative example that includes two correlated records belonging to \emph{Equity Underweight} and \emph{Equity Overweight} categories, respectively.
1)~ReDEE fails to extract the argument of \emph{EndDate}, \emph{Nov 14, 2018}, though it is successfully extracted as the argument of \emph{StartDate}.
It is because ReDEE extracts arguments of both roles using the same entity representations that are not role-aware.
% This limitation arises from GIT's extraction of arguments for both roles using the same entity representations, which lack role-awareness.
% In contrast, by extracting arguments of \emph{StartDate} and \emph{EndDate} based on respective role-aware representations, SEA succeed in extracting both arguments.
In contrast, SEA successfully extracts both arguments by leveraging role-aware representations for \emph{StartDate} and \emph{EndDate}, respectively.
2)~ReDEE fails to detect the \emph{Equity Overweight} event due to the interference of \emph{Equity Underweight} event and event-unrelated sentences.
However, SEA succeeds by detecting each event based on specific event type representation.

\textbf{Validity of Explicitly Aggregating.}
We sample 50 records per event type to verify the effectiveness of SEA.
We visualize the record-related Sentence Representations~(SRs) and Event Type Representations~(ETRs) of SEA via t-SNE ~\cite{van2008visualizing} in Figure \hyperlink{4}{4(a)}.
By explicitly aggregating, event type information is aggregated into respective ETR.
Thus, ETRs of the same type have a more compact distribution, and ETRs of different types are separated apart.
Detecting event on ETRs alleviates the interference suffered by the previous paradigm.
For each event type, we calculate the intra-class cosine similarity among 50 positive event detection features~(EDFs), and the inter-class similarity between 50 positive and 200 negative EDFs. 
Note that EDF indicates ETR for SEA.
As Figure \hyperlink{4}{4(b)} shows, SEA consistently has higher intra-class similarity and lower inter-class similarity than ReDEE.
It shows that positive ETRs are more similar, and keep farther distances with negative ETRs, proving detecting event on ETRs is easier than SRs of previous works.

\section{Conclusion}
In this paper, we propose the event Schema-based Explicitly Aggregating model~(SEA) for DEE.
SEA aggregates event information into event representations and decodes event records based on type-aware representations, thereby alleviating the interference suffered by the previous DEE paradigm.
Extensive experiments on two DEE datasets show that SEA significantly outperforms SOTA methods and demonstrate the effectiveness of our approach.

% References should be produced using the bibtex program from suitable
% BiBTeX files (here: strings, refs, manuals). The IEEEbib.bst bibliography
% style file from IEEE produces unsorted bibliography list.
% -------------------------------------------------------------------------
\bibliographystyle{IEEEbib}
\bibliography{strings,refs}

\end{document}